# Learning to Represent Mechanics via Long-term Extrapolation and Interpolation

Sébastien Ehrhardt[1]    Aron Monszpart[2]    Andrea Vedaldi[1]    Niloy Mitra[2]

[1] Department of Engineering Science, University of Oxford
{hyenal, vedaldi}@robots.ox.ac.uk
[2] Department of Computer Science, University College London
{a.monszpart, n.mitra}@cs.ucl.ac.uk

## Abstract

While the basic laws of Newtonian mechanics are well understood, explaining a physical scenario still requires manually modeling the problem with suitable equations and associated parameters. In order to adopt such models for artificial intelligence, researchers have handcrafted the relevant states, and then used neural networks to learn the state transitions using simulation runs as training data. Unfortunately, such approaches can be unsuitable for modeling complex real-world scenarios, where manually authoring relevant state spaces tend to be challenging. In this work, we investigate if neural networks can implicitly learn physical states of real-world mechanical processes only based on visual data, and thus enable long-term physical extrapolation. We develop a recurrent neural network architecture for this task and also characterize resultant uncertainties in the form of evolving variance estimates. We evaluate our setup to extrapolate motion of a rolling ball on bowl of varying shape and orientation using only images as input, and report competitive results with approaches that assume access to internal physics models and parameters.

## 1 Introduction

Animals can make remarkably accurate and fast predictions of physical phenomena in order to perform activities such as navigate, prey, or burrow. However, the nature of the mental models used to perform such predictions remains unclear and is still actively researched [9].

In contrast, science has developed an excellent *formal understanding* of physics; for example, mechanics is nearly perfectly described by Newtonian physics. While the constituent laws are simple and accurate, applying them to the description of a physical scenario is anything but trivial. First, the scenario needs to be *abstracted* (*e.g.*, by segmenting the scene into rigid objects, estimating physical parameters such as mass, linear and angular velocity, *etc.*, deciding which equations to apply, and so on). Then, prediction still requires the *numerical integration* of complex systems of equations. It is unlikely that this is the process of mental modeling followed by natural intelligences.

In an effort to develop model of physics that are more suitable for artificial intelligence, several authors have looked at the problem of learning physical predictors using deep neural networks. As a notable example, the recent Neural Physics Engine (NPE) [5] uses a neural network to learn the state transition function of mechanical systems. The state itself is handcrafted and includes physical parameters such as positions, velocities, and masses of rigid bodies. While this approach works well, a limitation is that it does not allow the network to learn its own abstraction of the physical system. This may prevent the model from learning efficient approximations of physics that are likely required to scale to complex real-world scenarios.

In this work, we ask whether a representation of the physical state of a mechanical system can be learned *implicitly* by a neural network, and whether this can be used to perform more accurate predictions. Compared to methods such as NPE, learning such a model is more challenging as no direct observations of the state of the system are available for training. Instead, the state is a hidden variable that must be inferred while solving a task for which supervision can be provided. As an example of such a task, we consider here the problem of *long-term physical extrapolation*.

Our approach to extrapolation is to develop a recurrent neural network architecture that not only contains an implicit representation of the state of the system, but is also able to evolve it through time. This differs from methods such as NPE that predict instantaneous variations of the system state, which are integrated in long-term predictions *a-posteriori*, after learning is complete. We show that accounting for the integration process during learning allows the network to learn an implicit representation of physics. Furthermore, we show that, in relatively complex physical setups, the resulting predictions can be competitive to a modified version of NPE, even when the inputs to the extrapolator are visual observations of the physical system instead of a direct knowledge of its initial state.

Since physical extrapolation is inherently ambiguous, we allow the model to explicitly estimate its prediction uncertainty by estimating the variance of a Gaussian observation model. We show that this modification further improves the quality of long-term predictions.

Empirically, we push our model by considering scenarios beyond the "flat" ones considered in most recent papers, such as objects sliding on planes and colliding, and look for the first time at the case of an object rolling on a non-trivial 3D shape, namely a bowl of varying shape and orientation, where both linear and angular momenta are tightly coupled.

As a final benefit of learning with long-term physical predictions, we show that our model is able, with minimal modifications, to learn not only to extrapolate physical trajectories, but also to interpolate them. Remarkably, interpolation is still obtained by computing the trajectory in a feed-forward manner, from the first to the last time step.

The rest of the paper is organized as follows. The relation of our work to the literature is discussed in section 2. The detailed structure of the proposed neural networks is given and motivated in section 3. These networks are extensively evaluated on a large dataset of simulated physical experiments in section 5. A summary of our finding can be found in section 6.

## 2 Related work

In this work we address the problem of long-term prediction from observation in a physical environment without voluntary perturbation, which is done by an implicit learning of physical laws. Our work is closely related to a range of recent works in the machine learning community.

**Learning intuitive physics.** To the best of our knowledge [4] was the first approach to tackle intuitive physics with the aim to answer a set of intuitive questions (*e.g.*, will it fall?) using physical simulations. Their simulations, however, used a sophisticated physics engine that incorporates prior knowledge about Newtonian physical equations. More recently [17] also used static images and a graphics rendering engine (Blender) to predict movements and directions of forces from a single RGB image. Motivated by the recent success of deep learning for image processing (*e.g.*, [12, 10]), they used a convolutional architecture to understand dynamics and forces acting behind the scenes from a static image and produced a "most likely motion" rendered from a graphics engine. In a different framework, [14] and [15] also used the power of deep learning to extract an abstract representation of the concept of stability of block towers purely from images. These approaches successfully demonstrated that not only was a network able to accurately predict the stability of the block tower but in addition, it could identify the source of the instability. Other approaches such as [2] or [7] also attempted to learn intuitive physics of objects through manipulation. These approaches, however, do not attempt to precisely model the evolution of the physical world.

**Learning dynamics.** Learning the evolution of an object's position also implies to learn about the object's dynamics regardless of any physical equations. While most successful techniques used LSTM-s [11], recent approaches show that propagation can also be done using a single cross-convolution kernel. The idea was further developed in [27] in order to generate a next possible image frame from a single static input image. The concept has been shown to have promising performance regarding longer term predictions on the moving MNIST dataset in [6]. The work of [19] also shows that an internal hidden state can be propagated through time using a simple deep recurrent architecture.

These results motivated us to propagate tensor based state representations instead of a single vector representation using a series of convolutions. Adversarial losses have also been used in [18] which shows good results in video segmentation. In the future we also aim to experiment with approaches inspired by [27].

**Learning physics.** The works of [26] and its extension [25] propose methods to learn physical properties of scenes and objects. However, in [26] the MCMC sampling based approach assumes the complete knowledge of the physical equations to estimate the correct physical parameters. In [25] deep learning has been used more extensively to replace the MCMC based sampling but this work also employs an explicit encoding and computation of physical laws to regress the output of their tracker. [22] also used physical laws to predict the movement of a pillow from unlabelled data though their approach was only applied to a fixed number of frames.

In another related approach [8] attempted to build an internal representation of the physical world. Using a billiard board with an external simulator they built a network which observing four frames and an applied force, was able to predict the 20 next object velocities. Generalization in this work was made using an LSTM in the intermediate representations. The process can be interpreted as iterative since frame generation is made to provide new inputs to the network. This can also be seen as a regularization process to avoid the internal representation of dynamics to decay over time which is different to our approach in which we try to build a stronger internal representation that will attempt to avoid such decay.

Other research attempted to abstract the physics engine enforcing the laws of physics as neural network models. [3] and [5] were able to produce accurate estimations of the next state of the world. Although the results look plausible and promising, reported results show in [5] that accurate long-term predictions are still difficult. Note, that their process is an iterative one as opposed to ours, which propagates an internal state of the world through time similarly to [20].

**Approximate physics with realistic output.** Other approaches also focused on learning to generate realistic future scenarios ([24] and [13]), or inferring collision parameters from monocular videos [16]. In these approaches the authors used physics based losses to produce visually plausible yet erroneous results. They however show promising results and constructed new losses taking into account additional physical parameters other than velocity. Note also that in [3] an energy-based loss has been used. It can be seen as a way to explicitly incorporate a knowledge of physics in the network while we aim to understand if we can make accurate prediction without explicit physics knowledge.

## 3 Method

In this section, we propose a new neural network model (see Fig. 1) that performs predictions in mechanical systems. Let $y_t$ be a vector of physical measurements taken at time $t$, such as the position of an object whose motion we would like to track. Physical systems satisfy a Markov condition, in the sense that there exists a state vector $h_t$ such that 1) measurements $y_t = g(h_t)$ can be predicted from the value of the state and 2) the state at the next time step $h_{t+1} = f(h_t)$ depends only on the current value of the state $h_t$. Uncertainty in the model can be encoded by means of transition $p(h_{t+1}|h_t)$ and observation $p(y_t|h_t)$ probabilities, resulting in a hidden Markov model.

Approaches such as NPE [5] start from an handcrafted definition of the state $h_t$. For instance, in order to model a scenario with two balls colliding, one may choose $h_t$ to contain the position and velocity of each ball. In this case, the observation function $g$ may be as simple as extracting the position components from the state vector. The goal of NPE is then to learn a neural network approximator $\phi$ of the transition function $f$. In practice, the authors of [5] suggest that it is often easier to predict a rate of change $\Delta_t$ for some of the physical parameters (*e.g.* the balls' velocities), which can then be integrated to update the state: $h_{t+1} = \tilde{f}(h_t, \Delta_t)$ where $\tilde{f}$ is an hand-crafted integrator and the neural predictor estimates the change $\Delta_t = \phi(h_t)$.

While these approaches have several advantages [5], there are several limitations too. First, approaches such as NPE require to handcraft the state representation $h_t$. Even in the simple case of the colliding balls, the choice of state is ambiguous; for example, one could include in the state the radius, mass and elasticity and friction coefficients. In more complex situations, choosing a good state representation may be rather difficult. Overall, this choice is best left to learning. Second, the state must be observable during training in order to learn the transition function. Third, learning does not account for the effect of accumulating errors through integration as integration is applied only after learning. Finally, the initial value of the state $h_0$ must be known in order to initialize the predictor, whereas in many applications one would like to start from *sensory inputs* $x_t$ such as images of the physical system [8].

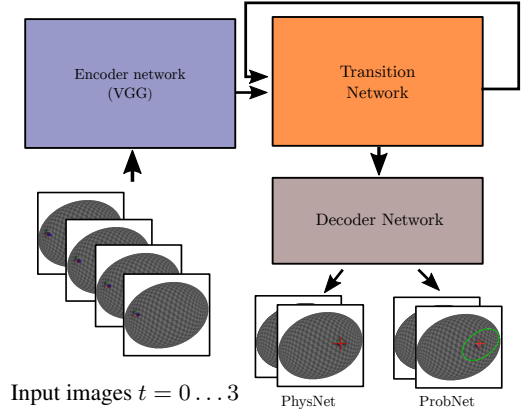

Figure 1: Overview of our proposed pipeline. The first four images of a sequence first pass through a partially pre-trained feature encoder network to build the concept of physical state. It then recursively passes through a transition layer to produce long-term predictions about the future states of the objects. It is then decoded to produce state estimates. While our *PhysNet* model is trained to regress the next states, the *ProbNet* model trained with the log-likelihood loss is also able to handle the notion of uncertainty thanks to its extended state space.

We propose here an approach to address these difficulties. We assume that the state $h_t$ is a *hidden variable*, to be determined as part of the learning process. Since the $h_t$ cannot be observed, the transition function $h_{t+1} = f(h_t)$ cannot be estimated directly as in the NPE. Instead, it must be inferred as a *good explanation* of the physical measurements $y_t$. Since the evolution of the state $h_t$ cannot be learned by observing measurement $y_t$ in isolation, we supervise the system by explaining sequences $\mathbf{y}_{[0,T)} = (y_0, \ldots, y_{T-1})$. This requires to move the integration step *inside the network*, which we do by mean of a recurrent neural network architecture. This has the added advantage of making learning aware of the integration process, which helps improving accuracy.

The model is analogous to a Hidden Markov Model. Recall that such models are often learned by maximizing the likelihood of the observations after marginalizing the hidden state.[1] However, since we are interested in extrapolating future observations from past ones, we consider instead long-term extrapolation as supervisory signal. In order to do so we learn: 1) a transition function $h_{t+1} = \phi(h_t)$ that evolves the state through time, 2) a decoder function that maps the state $h_t$ to an observation $y_t = \phi_{\text{dec}}(h_t)$, and 3) an encoder function that estimates the state $h_t = \phi_{\text{enc}}(\mathbf{x}_{(t-T_0,t]})$ from the $T_0$ most recent sensor readings (alternatively $h_t = \phi_{\text{enc}}(\mathbf{y}_{(t-T_0,t]})$ can use the $T_0$ most recent observations).

In the experiments (section 5) we will show that the added flexibility of learning an internal state representation automatically can still provide a good prediction accuracy even when the complexity of the physical scenarios increases. The rest of the section discusses the three modules, encoder, transition, and decoder maps, as well as the loss function used for training. Further technical details can be found in section 5.

**(i) Encoder map: from images to state.** The goal of the encoder map is to take $T_0$ consecutive video frames recording the beginning of the object motion and to produce an estimate $h_0 = \phi_{\text{enc}}(\mathbf{x}_{(-T_0,0]})$

---

[1] Formally, the Markov model is given by $p(\mathbf{y}_{[0,T)]}, \mathbf{h}_{[0,T)}) = p(h_0)p(y_0|h_0) \prod_{t=0}^{T-2} p(h_{t+1}|h_t)p(y_{t+1}|h_{t+1})$; traditionally, $p$ can be learned as the maximizer of the log-likelihood $\max_p E_y[\log E_\mathbf{h}[p(\mathbf{y}, \mathbf{h})]]$, where we dropped the subscripts for compactness. Learning to interpolate/extrapolate can be done by considering subsets $\bar{\mathbf{y}} \subset \mathbf{y}$ of the measurements as given and optimizing the likelihood of the conditional probability $\max_p E_\mathbf{y}[\log E_\mathbf{h}[p(\mathbf{y}, \mathbf{h}|\bar{\mathbf{y}})]]$.

of the initial state of the physical system. In order to build this encoder, we follow [8] and concatenate the RGB channels of the $T_0$ images in a single $H_i \times W_i \times 3T_0$. The latter is passed to a convolutional neural network $\phi_{\text{enc}}$ outputting a feature tensor $s_0 \in \mathbb{R}^{H \times W \times C}$, used as internal representation of the system. We also add to the state a vector $p_0 \in \mathbb{R}^2$ to store the 2D projection of the object location on the image plane, so that $h_t = (s_t, p_t)$.

**(ii) Transition map: evolving the state.** The state $h_t$ is evolved through time by learning the transition function $\phi : h_t \mapsto h_{t+1}$, where $h_0$ is obtained from the encoder map, so that $h_t = \phi^t(\phi_{\text{enc}}(\mathbf{x}_{(-T_0, 0]}))$. The state $s_t$ is updated by using a convolutional network $s_{t+1} = \phi_s(s_t)$ whereas $p_t$ is updated incrementally as $p_{t+1} = p_t + \phi_p(s_t)$, where $\phi_p(s_t)$ is estimated using a single layer perceptron regressor. Hence $(s_{t+1}, p_{t+1}) = \phi(s_t, p_t) = (\phi_s(s_t), p_t + \phi_p(s_t))$. We found that explicitly incorporating an additive update significantly improves the performance of the model.

**(iii) Decoder map: from state to probabilistic predictions.** Since we added for convenience the projected object position $p_t$ to the state, the decoder map $\hat{y}_t = \phi_{\text{dec}}(s_t, p_t) = p_t$ simply extracts and returns that part of the state. Training optimizes the average $L^2$ distance between ground truth $y_t$ and predicted $\hat{y}_t$ positions $\frac{1}{T}\sum_{t=0}^{T-1} \|\hat{y}_t - y_t\|^2$.

Since extrapolation is inherently ambiguous, the $L^2$ prediction error increases with time, which may unbalance learning. In order to address this issue, we allow the model to explicitly and dynamically express its prediction uncertainty by outputting the mean and variance $(\mu_t, \Sigma_t)$ of a bivariate Gaussian observation model. The $L^2$ loss is then replaced with the negative log likelihood $-\frac{1}{T}\sum_{t=0}^{T-1} \log \mathcal{N}(y_t; \mu_t, \Sigma_t)$.

In order to estimate $\mu_t$ and $\Sigma_t$, the incremental state component $p_t = (\mu_t, \lambda_{1,t}, \lambda_{2,t}, \theta_t)$ is extended to include both the mean as well as the eigenvalues and rotation of the variance matrix $\Sigma_t = R(\theta_t)^\top \text{diag}(\lambda_{1,t}, \lambda_{2,t}) R(\theta_t)$. In order to ensure numerical stability, eigenvalues are constrained to be in the range $[0.01 \ldots 100]$ by setting them as the output of a scaled and translated sigmoid $\lambda_{i,t} = \sigma_{\lambda,\alpha}(\beta_{i,t})$, where $\sigma_{\lambda,\alpha}(z) = \lambda/(1 + \exp(-z)) + \alpha$.

## 4 Experimental setup

In our experimental setup (Fig. 2), we consider a sphere rolling inside a 3D (bowl) surface. When the bowl is a hemisphere we refer to the setup as 'Bowl,' and in the more general case as 'Ellipse' (see Table 1).

We use $\mathbf{p} = (p_x, p_y, p_z) \in \mathbb{R}^3$ to denote a point in 3D space or a vector (direction). The camera center is placed at location $(0, 0, c_z)$, $c_z > 0$ and looks downward along vector $(0, 0, -1)$ using orthographic projection such that the point $(p_x, p_y, p_z)$ projects to a pixel $(p_x, p_y)$ in the image.

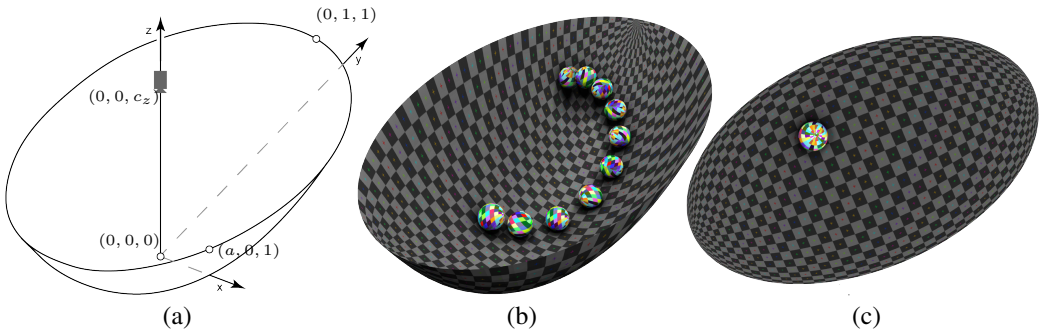

Figure 2: We consider the problem of understanding and extrapolating mechanical phenomena with recurrent deep networks. (a) Experimental setup: an orthographic camera looks at a ball rolling in a 3D bowl. (b) Example of a 3D trajectory in the 3D bowl simulated using Blender 2.77's OpenGL renderer. (c) An example of a rendered frame in the 'Ellipse' experiment that is fed to our model as input.

We model the bowl as the bottom half of an ellipsoid given by $x^2/a^2 + y^2 + (z-1)^2 = 1$ with its axes aligned to the XYZ axes and its bottom point being at the origin. We vary the ellipsoid shape by sampling $a \in U[0.5, 1]$ for the 'Ellipse' case and setting $a = 1$ (*i.e.*, a hemisphere) for the 'Bowl' case. The bowl is given a checker board pattern. Finally, the bowl is given a random rotation $\gamma \in U[-\pi/2, \pi/2]$ only about the z-axis to randomly orient it.

We consider a rolling object in the form of a ball with radius $\rho = 0.04$ with its center of mass at time $t$ being located at $\mathbf{q}^t = (q_x^t, q_y^t, q_z^t)$, so that its center of mass is imaged at pixel $(q_x^t, q_y^t)$ at any time $t$. The ball has a fixed color texture attached to its surface, so it appears as a painted object. We initially position the ball at angles $(\theta, \phi)$ with respect to the the bowl center, where the elevation $\theta$ is uniformly sampled in the range $\theta \in U[-9\pi/10, -\pi/2]$ and the azimuth $\phi \in U[-\pi, \pi]$. We set the minimum elevation to $-9\pi/10$ to avoid starting the ball at the bottom of the bowl. In the end, the ball will be resting on the bowl surface. The ball is either textured with random color patches or uniformly colored in white in order to study the impact of observing the ball rotation.

We set the initial orientation of the ball by uniformly sampling its xyz Euler angles in $[-\pi, \pi]$. We set its initial velocity $\mathbf{v}$ by first sampling $v_x, v_y$ uniformly in the range $[5, 10]$, assigning each of $v_x, v_y$ a random sign, and then projecting vector $(v_x, v_y, 0)$ so that the resulting velocity vector is *tangential* to the underlying supporting bowl.

Note that, while several parameters of the ball state are included in the *observation* vector $\mathbf{y}^\alpha_{[-T_0, T)}$, these are *not* part of the state of the neural network, which is inferred automatically. The network itself is tasked with *predicting* part of these measurements, but their meaning is not hardcoded.

**Simulation setup.** For efficiency, we extract multiple sub-sequences $\mathbf{x}^\alpha_{[-T_0, T)}$ form a single longer simulation (training, test, and validation sets are completely independent). The simulator runs at 120fps for accuracy, but the data is subsampled to 40fps. We use Blender 2.77's OpenGL renderer and the Blender Game Engine (relying on Bullet 2 as physics engine). The ball is a textured sphere with unit mass. We found that changing the friction parameter of the bowl or the ball does not influence the motion. Therefore, we added translation and rotation damping (both set to 0.1 in Blender) to the sphere's animation properties in order to simulate energy loss due to friction. The simulation parameters were set as: max physics steps = 5, physics substeps = 3, max logic steps = 5, FPS = 120. Rendering used white environment lighting (energy = 0.7) and no other light source. The object color was set to a colored checkerboard texture in order to enable the visual perception of rotation. We used 70% the data for training, 15% for validation, and 15% for test. During training we start observation at a random time while it is fixed for test. The output images were stored as $128 \times 128$ color JPEG files.

## 5 Experiments

### 5.1 Baselines

**Least squares fit.** We compare the performance of our methods to two simple least squares baselines: Linear and Quadratic. In both cases we fit two least squares polynomials to the screen-space coordinates of the first $T = 10$ frames, which are not computed but rather given as inputs. The polynomials are of first and second degree(s), respectively. Note, that being able to observe the first 10 frames is a large advantage compared to the networks, which only see the first $T_0 = 4$ frames.

**NPE.** NPE [5] training was done using available online code. We used the same training procedure as reported in [5]. NPE++ additionally takes angle and angular velocities as parameters and also predicts angular velocity. In the case of the elliptic bowl, both scaling and bowl rotation angle are given as input to the networks. In this case NPEs methods carry forward the estimated states via the network.

While the previously mentioned methods start from state inputs, note that our models work with raw images as direct observation of the world. Physical properties are then deduced from the observation and then integrated through our Markov model. Thus we do not need a simulator to estimate parameters of the physical worlds (such as scaling and rotation angle in the NPE case) and can train our model on changing environment without requiring additional external measurements of the underlying 3D spaces.

## 5.2 Results

**Implementation details.** The encoder network $\phi_{\text{enc}}$ is obtained by taking the ImageNet-pretrained VGG16 network [21] and retaining the layers up to conv5 (for an input image of size $(H_i, W_i) = (128, 128, 3)$ this results in a $(8, 8, 512)$ state tensor $s_t$). The filter weights of all layers except conv1 are retained for fine-tuning on our problem. conv1 is reinitialized as filters must operate on images with $3T_0$ channels. The transition network $\phi_s(s_t)$ uses a simple chain of two convolution layers (with 256 and 512 filters respectively, of size $3 \times 3$, stride 1, and padding 1 interleaved by a ReLU layer. Network weights are initialized by sampling from a Gaussian distribution.

Training uses a batch size of 50 using the first 20 or 40 positions (and angular velocity when explicitly mentioned) of each video sequence using RMSProp [23]. In our methods we start with a learning rate of $10^{-5}$ and decrease learning rates by a factor of 10 when no improvements of the $L_2$ position loss have been found after 100 consecutive epochs. Training is halted when the $L_2$ loss hasn't decreased after 200 successive epochs; 2,000 epochs were found to be usually sufficient for convergence. Note here than in every cases where we estimate the angular velocity the corresponding $L_2$ loss on the latter is simply added to the network's existing loss.

Since during the initial phases of training the network is very uncertain, the model using the Gaussian log-likelihood loss was found to get stuck on solutions with very high variance $\Sigma(t)$; to solve this issue, the regularizer $\lambda \sum_t \det \Sigma(t)$ was added to the loss, setting $\lambda = 0.01$.

In all our experiments we used Tensorflow [1] r0.12 on a single NVIDIA Titan X GPU. In the following we will refer to our model that has been trained optimizing on the $L^2$ loss over positions as *PhysNet* otherwise *ProbNet* for log-likelihood loss. When also predicting angular velocities the suffix '++' is added to the name of the model.

Table 1: **Long term predictions.** The *PhysNet* and *ProbNet* models observed the $T_0 = 4$ first frames as input. *PhysNet* ++ and *ProbNet* ++ additionally estimate angular velocity at each time step adding a $L^2$ angular velocity loss to our position loss. All networks have been trained to predict the $T = 20$ first positions, except for the NPE and NPE++ which were given $T_0 = 4$ states as input and train to predict state at time $T_0 + 1$. We report here results for time $T = 20$ and $T = 40$. For each time we report on the left $L^2$ position loss and $L^2$ angular loss on the right. Perplexity ($\log_e$ values shown in the table) is defined as $2^{-\mathbb{E}[\log_2(p(x))]}$ where $p$ is the estimated posterior distribution.

| Method | Images | Bowl $L^2$ (Perplexity) 20 | | 40 | | Ellipse $L^2$ (Perplexity) 20 | | 40 | | Ellipse (no ball texture) $L^2$ (Perplexity) 20 | | 40 | |
|---|---|---|---|---|---|---|---|---|---|---|---|---|---|
| Linear | No | 39.2 | 7.5 | 127.5 | 17.9 | 61.9 | 23.3 | 20.1 | – | – | – | – | – |
| Quadratic | No | 164.3 | 18.4 | 120.1 | 861.2 | 11.7 | 14.8 | 93.1 | 70.6 | – | – | – | – |
| NPE | No | **2.6** | – | **6.0** | – | 3.2 | – | **6.1** | – | – | – | – | – |
| NPE++ | No | 2.9 | **0.8** | 8.1 | **2.0** | 5.7 | 1.7 | 17.5 | **3.1** | – | – | – | – |
| *PhysNet* | Yes | 3.0 | – | 29.7 | – | 2.5 | – | 20.6 | | 5.2 | – | 44.6 | – |
| *PhysNet* ++ | Yes | 3.5 | 15.9 | 15.9 | 11.9 | **2.1** | **1.0** | 16.1 | 4.4 | 1.6 | **1.0** | 16.2 | 3.8 |
| *ProbNet* | Yes | 2.9 | – | 24.2 | – | 2.9 | – | 21.8 | – | 3.1 | – | 24.0 | – |
| | | (4.5) | | (21.9) | | (32.1) | | (54.0) | | (5.0) | | (12.7) | |
| *ProbNet* ++ | Yes | 3.4 | 1.2 | 15.3 | 3.4 | 4.0 | 1.8 | 16.7 | 3.8 | 4.3 | 1.3 | 15.0 | 3.5 |
| | | (4.7) | | (9.2) | | (4.5) | | (9.3) | | (4.5) | | (8.2) | |

**Extrapolation.** Table 1 and Fig. 3 compare the baseline predictors and the four networks on the task of long term prediction of the object trajectory. All methods observed only the first $T_0 = 4$ inputs (whether frames or object states) except for the linear and quadratic baselines, and aimed to extrapolate the trajectory to 40 time steps. In that sense predictions can be seen as "long terms" relative to the number of inputs.

Table 1 reports the average $L^2$ errors at time $T_{\text{test}} = 20$ and 40 for the different estimated parameters. However all of the methods can perform arbitrary long predictions. In particular our methods are only trained to predict 20 first positions and reveal to be still competitive with NPEs methods at $T = 40$.

Our networks perform reasonably well compared to the NPEs methods using only images as inputs. In our scenario we can see the different NPEs as upper limits of our experiments since they do have access to the state space (complete or not). Furthermore adding angular velocity has shown to improve performances of our models while it decreases the accuracy of the NPE++ predictions in that case. Besides probability based losses show that our models were also able to predict uncertainty in its outputs even in the case of unobserved scenarios.

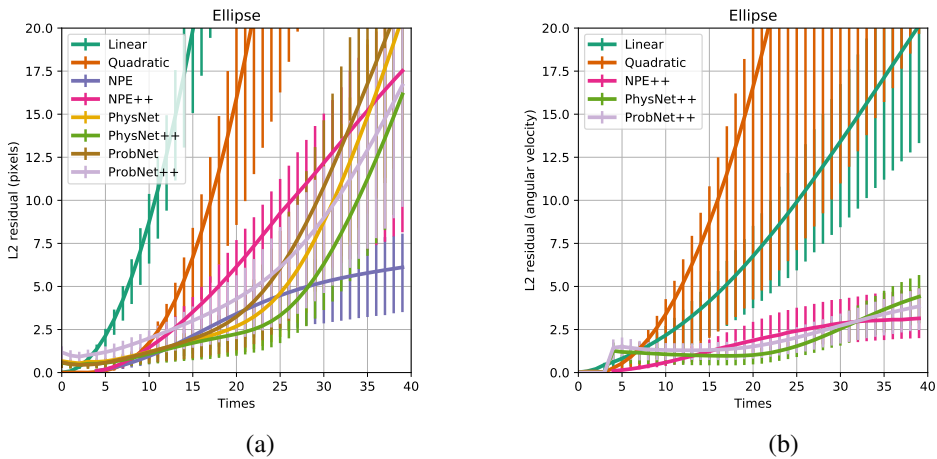

(a)                                                   (b)

Figure 3: **Quantitative results.** error evolution on Ellipse experiments for all time steps up to 40. Error bars denote $25^{th}$ and $75^{th}$ percentiles of the $L^2$ loss. (a) $L^2$ position loss in pixels. (b) $L^2$ angular velocity loss.

In addition training on a dataset where angular velocity was not explicitly seen (Ellipse no-texture in Table 1) shows that our models can still provide encouraging results in that case. It managed to accurately deduce angular velocity without seeing the ball spinning.

**Interpolation.** In order to remove ambiguity of a short term observed motion one can just indicates the final desired state. In this experiment we concatenate to the first $T_0 = 4$ input frames the last frame observed at $T = 40$ and give it as an input to a model with the same architecture as *PhysNet*. This model was then trained using the same aforementioned method with the only difference that we also extract last positions from the first extracted feature. While this idea is fairly simple it shows to be very efficient in practice as shown in Table 2 as it efficiently removed the motion ambiguity.

Table 2: **Interpolation.** *InterpNet* is essentially the same as *PhysNet* but takes as input a concatenation of first $T_0 = 4$ frames and last frame at $T = 40$. All networks have been trained to predict the $T = 40$ first positions. *InterpNet* predicts $T = 40$ positions using the first extracted feature.

| Method | Bowl $L^2$ | | | | Ellipse $L^2$ | | | |
|---|---|---|---|---|---|---|---|---|
| | 10 | 20 | 30 | 40 | 10 | 20 | 30 | 40 |
| *InterpNet* | 1.37 | 1.84 | 1.65 | 1.02 | 1.03 | 1.60 | 1.35 | 0.65 |
| *PhysNet* | 2.19 | 3.64 | 3.94 | 4.99 | 1.40 | 2.39 | 2.71 | 3.00 |

# 6 Conclusions

In this paper we studied the possibility of abstracting the knowledge of physics using a single neural network with a recurrent architecture for long term predictions. We compared our model to strong baselines on the non-trivial motion of a ball rolling on a 3D bowl with different possible shapes. As opposed to other concurrent approaches we do *not* integrate physical quantities but implicitly encode the states in a feature vector that we can propagate through time.

Our experiments on synthetic simulations indicate that we can still make reasonable predictions without requiring an explicit encoding of the state space. Besides they are also able to estimate a distribution over such parameters to account for uncertainty in the predictions. While keeping the same architecture we also show that we were able to remove motion ambiguity by showing the network its targeted final states. However, the internal state propagation mechanism is still limited by its ability to make accurate long term predictions outside observed regimes.

In the future we will aim to bring more robustness to our models by enforcing invariance to observed regimes to enable longer accurate predictions. Besides, a next obvious step will also be to test the framework on video footage obtained from real-world data in order to assess the ability to do so from visual data affected by real nuisance factors.